\documentclass[10pt,twocolumn,letterpaper]{article}

\usepackage{cvpr}              

\usepackage{graphicx}
\usepackage{amsmath}
\usepackage{amssymb}
\usepackage{booktabs}
\usepackage{multirow}
\usepackage{adjustbox}
\usepackage[accsupp]{axessibility}
\usepackage{xcolor}
\usepackage{float}
\usepackage{makecell}
\usepackage[pagebackref,breaklinks,colorlinks]{hyperref}

\usepackage{pifont}
\newcommand{\xmark}{\ding{55}}%

\usepackage[capitalize]{cleveref}
\crefname{section}{Sec.}{Secs.}
\Crefname{section}{Section}{Sections}
\Crefname{table}{Table}{Tables}
\crefname{table}{Tab.}{Tabs.}

\def\cvprPaperID{4641} 
\def\confName{CVPR}
\def\confYear{2023}

\newcommand{\dname}{Meta Omnium}

\begin{document}

\title{Meta Omnium: A Benchmark for General-Purpose Learning-to-Learn}

\author{Ondrej Bohdal$^{1,*}$, Yinbing Tian$^{2,*}$, Yongshuo Zong$^{1}$, Ruchika Chavhan$^{1}$,\\ Da Li$^{3}$, Henry Gouk$^{1}$, Li Guo$^{2}$, Timothy Hospedales$^{1,3}$\\[1.5mm]
$^{1}$\fontsize{10pt}{\baselineskip}\selectfont The University of Edinburgh \ $^{2}$\fontsize{10pt}{\baselineskip}\selectfont Beijing University of Posts and Telecommunications \\
$^{3}$\fontsize{10pt}{\baselineskip}\selectfont Samsung AI Center, Cambridge $^{*}$\fontsize{10pt}{\baselineskip}\selectfont Joint-First Authors\\
}

\maketitle

\begin{abstract}
   Meta-learning and other approaches to few-shot learning are widely studied for image recognition, and are increasingly applied to other vision tasks such as pose estimation and dense prediction. This naturally raises the question of whether there is any few-shot meta-learning algorithm capable of generalizing across these diverse task types? To support the community in answering this question, we introduce \dname{}, a dataset-of-datasets spanning multiple vision tasks including recognition, keypoint localization, semantic segmentation and regression. We experiment with popular few-shot meta-learning baselines and analyze their ability to generalize across tasks and to transfer knowledge between them. \dname{} enables meta-learning researchers to evaluate model generalization to a much wider array of tasks than previously possible, and provides a single framework for evaluating meta-learners across a wide suite of vision applications in a consistent manner. The code and dataset can be accessed from the project page at \url{https://edi-meta-learning.github.io/meta-omnium}.
\end{abstract}

\section{Introduction}
\label{sec:intro} 
Meta-learning is a long-standing research area that aims to replicate the human ability to learn from a few examples by learning-to-learn from a large number of learning problems \cite{thrun1998learntolearn}. This area has become increasingly important recently, as a paradigm with the potential to break the data bottleneck of traditional supervised learning \cite{hospedales20201metaSurveyPAMI,wang2020fslSurvey}. While the largest body of work is applied to image recognition, few-shot learning algorithms have now been studied in most corners of computer vision, from semantic segmentation \cite{li2020fss1000} to pose estimation \cite{patacchiola2020bayesian} and beyond. Nevertheless, most of these applications of few-shot learning are advancing independently, with increasingly divergent application-specific methods and benchmarks. This makes it hard to evaluate whether few-shot meta-learners can solve diverse vision tasks. Importantly it also discourages the development of meta-learners with the ability to learn-to-learn across tasks, transferring knowledge from, e.g., keypoint localization to segmentation -- a capability that would be highly valuable for vision systems if achieved. 

\begin{figure}[t]
  \centering
   \includegraphics[width=1.0\linewidth]{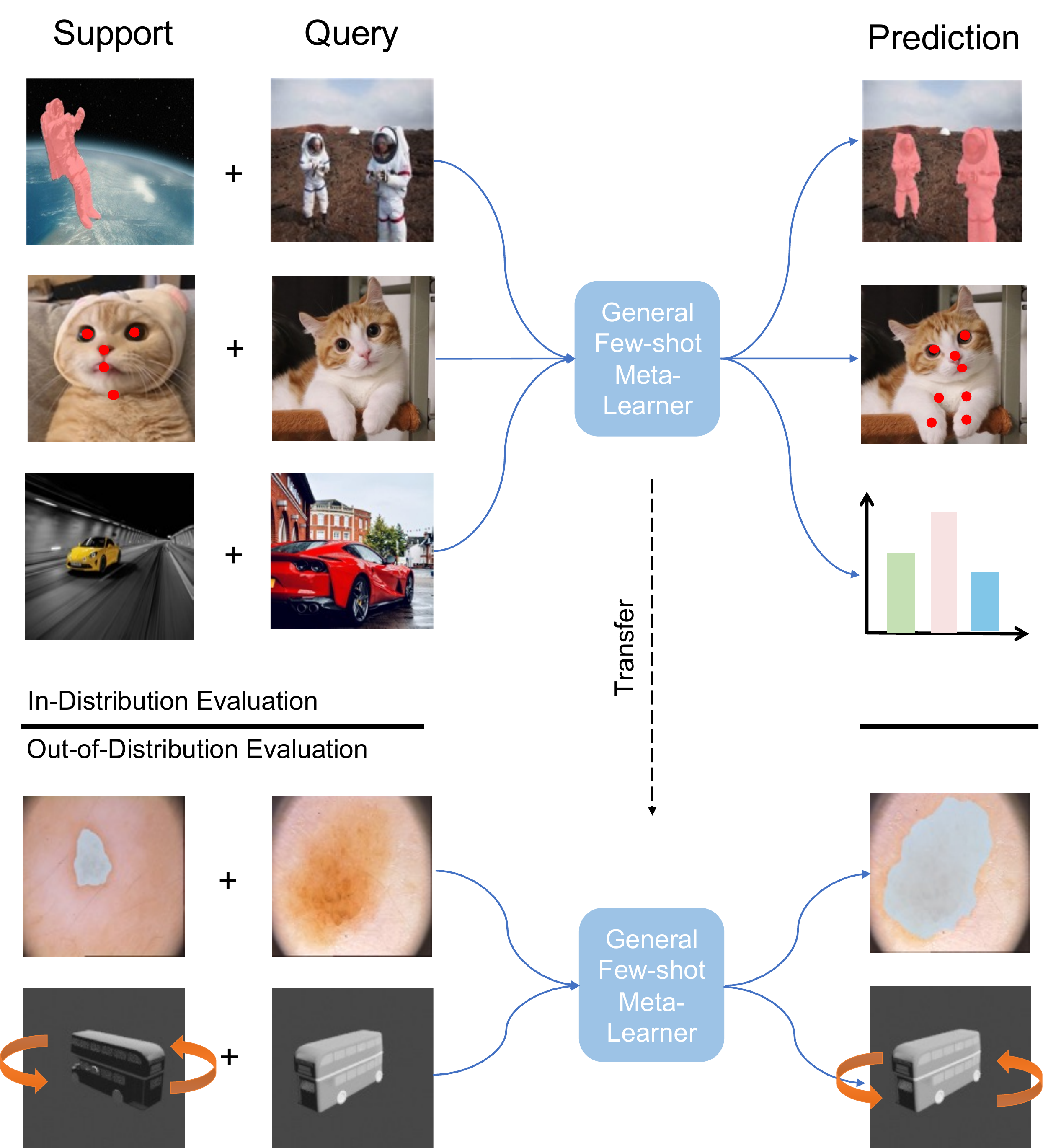}
   \caption{Illustration of the diverse visual domains and task types in \dname{}. Meta-learners are required to generalize across multiple task types, multiple datasets, and held-out datasets.}
   \label{fig:teaser}
\end{figure}

\begin{table*}[t]
\resizebox{1.0\textwidth}{!}{
\begin{tabular}{lrrllllll}
\hline\textbf{Dataset}    & \multicolumn{1}{l}{\textbf{Num Tasks}} & \multicolumn{1}{l}{\textbf{Num Domains}} & \textbf{Num Imgs} & \textbf{Categories}      & \textbf{Size} & \textbf{Lightweight} & \textbf{Multi-Task} & \textbf{Multi-Domain} \\
\hline
Omniglot \cite{lake2015ppi}           & 1                                      & 1                                        & 32K               & 1623 & 148MB                 & \checkmark                    & \xmark                   & \xmark                     \\
miniImageNet \cite{vinyals2016oneShot}       & 1                                      & 1                                        & 60K               & 100  & 1GB                   & \checkmark                    & \xmark                   & \xmark                     \\
Meta-Dataset \cite{triantafillou2020metaDataset}       & 1                                      & 7$\sim$10                                     & 53M               &     43$\sim$1500                     & 210GB                 & \xmark                    & \xmark                   & \checkmark                     \\
VTAB  \cite{zhai2019largeVTAB}              & 1                                      & {3$\sim$19}                 & 2.2M              &    2$\sim$397                      & 100GB                 & \xmark                    & \xmark                   & \checkmark                        \\
FSS1000  \cite{li2020fss1000}              & 1                                      & 1                 & 10000              &      1000   &        670MB          & \checkmark                    & \xmark                   & \xmark                        \\
Meta-Album \cite{ullah2022metaAlbum}         & 1                                      & {10$\sim$40}                & 1.5M              &   19$\sim$706                       & 15GB                  & \checkmark                       & \xmark                   & \checkmark                        \\

\hline
\dname{}      & 4                                      & 21                                    &    160K               &                     2$\sim$706     &        3.1GB             & \checkmark                    & \checkmark                   & \checkmark                  \\  \hline
\end{tabular}}
\caption{Feature comparison between \dname{} and other few-shot meta-learning benchmarks. \dname{} uniquely combines a rich set of tasks and visual domains with a lightweight size for accessible use. }\label{tab:comparison}
\end{table*}

The overall trend in computer vision \cite{ghiasi2021multiTask,radford2021learningCLIP} and AI \cite{reed2022gatoAgent,baevski2022data2vec} more generally is towards more general-purpose models and algorithms that support many tasks and ideally leverage synergies across them. However, it has not yet been possible to explore this trend in meta-learning, due to the lack of few-shot benchmarks spanning multiple tasks. State-of-the-art benchmarks \cite{triantafillou2020metaDataset,ullah2022metaAlbum} for visual few-shot learning are restricted to image recognition across a handful of visual domains. There is no few-shot benchmark that poses the more substantial challenge \cite{yu2020gradientSurgery,ruder2017mtlSurvey} of generalizing across different \emph{tasks}. We remark that the term \emph{task} is used differently in few-shot meta-learning literature \cite{finn2017maml,wang2020fslSurvey,hospedales20201metaSurveyPAMI} (to mean different image recognition problems, such as cat vs dog or car vs bus) and the multi-task literature \cite{yu2020gradientSurgery,ruder2017mtlSurvey,ghiasi2021multiTask,yang2015mdmtl} (to mean different kinds of image understanding problems, such as classification vs segmentation). In this paper, we will use the term \emph{task} in the multi-task literature sense, and the term \emph{episode} to refer to tasks in the meta-learning literature sense, corresponding to a support and query set. 

We introduce \dname{}, a dataset-of-datasets spanning multiple vision tasks including recognition, semantic segmentation, keypoint localization/pose estimation, and regression as illustrated in Figure~\ref{fig:teaser}. Specifically, \dname{} provides the following important contributions: (1) Existing benchmarks only test the ability of meta-learners to learn-to-learn within tasks such as classification \cite{triantafillou2020metaDataset,ullah2022metaAlbum}, or dense prediction \cite{li2020fss1000}. \dname{} uniquely tests the ability of meta-learners to learn across multiple task types. (2) \dname{} covers multiple visual domains (from natural to medical and industrial images). (3) \dname{} provides the ability to thoroughly evaluate both in-distribution and out-of-distribution generalisation. (4) \dname{} has a clear hyper-parameter tuning (HPO) and model selection protocol, to facilitate future fair comparison across current and future meta-learning algorithms.
(5), Unlike popular predecessors, \cite{triantafillou2020metaDataset}, and despite the diversity of tasks, \dname{} has been carefully designed to be of moderate computational cost, making it accessible for research in modestly-resourced universities as well as large institutions.
Table~\ref{tab:comparison} compares \dname{} to other relevant meta-learning datasets.

We expect \dname{} to advance the field by encouraging the development of meta-learning algorithms capable of knowledge transfer across different tasks -- as well as across learning episodes within individual tasks as is popularly studied today \cite{finn2017maml,wang2020fslSurvey}. In this regard, it provides the next step of the level of a currently topical challenge of dealing with heterogeneity in meta-learning \cite{triantafillou2020metaDataset,vuorio2019multimodalMAML,abdollahzadeh2021revisitMMML,li2021metaCoding}. While existing benchmarks have tested \emph{multi-domain} heterogeneity (e.g., recognition of written characters and plants within a single network) \cite{triantafillou2020metaDataset,ullah2022metaAlbum} and shown it to be challenging, \dname{} tests \emph{multi-task learning} (e.g., character recognition vs plant segmentation). This is substantially more ambitious when considered from the perspective of common representation learning. For example, a representation tuned for recognition might benefit from rotation \emph{invariance}, while one tuned for segmentation might benefit from rotation \emph{equivariance} \cite{ericsson2021whySSRL,xiao2021whatContrastive,chavhan2023amortised}. Thus, in contrast to conventional within-task meta-learning benchmarks that have been criticized as relying more on common representation learning than learning-to-learn \cite{tian2020rethinking,raghu2020rapidANIL}, \dname{} better tests the ability of learning-to-learn since the constituent tasks require more diverse representations.

\section{Related Work}

\subsection{Meta-learning Benchmarks}

The classic datasets in few-shot meta-learning for computer vision are Omniglot \cite{lake2015ppi} and miniImageNet \cite{vinyals2016oneShot}. Later work criticized these for having insufficient task (episode) diversity and tieredImageNet \cite{ren2018metaSSL} used the class hierarchy of ImageNet to enforce more diversity between meta-train and meta-test episodes. The main  contemporary benchmarks are CD-FSL \cite{guo2020broader}, which challenges few-shot learners to generalize to new visual domains; and Meta-Dataset  \cite{triantafillou2020metaDataset} and Meta-Album \cite{ullah2022metaAlbum}, which go further in requiring few-shot learners to learn from a mixture of visual domains. Such multi-domain heterogeneous meta-learning turns out to be challenging. A related benchmark to Meta-Dataset is VTAB \cite{zhai2019largeVTAB}, which similarly provides multiple domains for evaluating  data-efficient visual recognition, but their focus is on evaluating representation transfer from large-scale pre-training rather than learning-to-learn and meta-learning. VTAB+MD \cite{dumoulin2021a} compare representation transfer and meta-learning approaches on the Meta-Dataset tasks. However, none of these benchmarks address multi-task meta-learning as considered here (Figure~\ref{fig:teaser}). 

Outside of recognition, task-specific few-shot benchmarks have been proposed in vision problems of semantic segmentation \cite{li2020fss1000}, regression \cite{gao2022metaRegression}, pose/keypoint estimation \cite{xu2022poseForAll}, etc. These are mostly slightly behind the complexity of the recognition benchmarks with regards to being single-domain, with the  exception of \cite{xu2022poseForAll}. With regard to multi-task meta-learning as considered here, the only existing benchmark is meta-world \cite{Yu2019}, which is specific to robotics and reinforcement learning rather than vision.

We also mention taskonomy \cite{zamir2018taskonomy} as a popular dataset that has been used for multi-task learning. However, it is not widely used for few-shot meta-learning. This is because, although taskonomy has many tasks, unlike the main meta-learning benchmarks \cite{triantafillou2020metaDataset,li2020fss1000}, there are not enough visual concepts within each task to provide a large number of concepts for meta-training and a disjoint set of concepts to evaluate few-shot learning for meta-validation and meta-testing. 

\subsection{Heterogeneity in Meta-Learning}
There are several sophisticated methods in the literature that highlighted the challenge of the task to address heterogeneity in meta-learning. These have gone under various names such as multi-modal meta-learning \cite{vuorio2019multimodalMAML,abdollahzadeh2021revisitMMML,liu2021multi} -- in the sense of multi-modal probability distribution (over tasks/episodes). However, with the exception of \cite{liu2021multi}, these have mostly not been shown to scale to the main multi-modal benchmarks such as Meta-Dataset \cite{triantafillou2020metaDataset}. A more common approach to achieving high performance across multiple  heterogeneous domains such as those in Meta-Dataset is to train an ensemble of feature extractors across available training domains, and fuse them during meta-testing \cite{dvornik2020selecting,li2021universalFSL,li2021universalFSL}. However, this obviously incurs a substantial additional cost of maintaining a model ensemble. 
In our evaluation, we focus on the simpler meta-learners that have been shown to work in challenging multi-domain learning scenarios \cite{triantafillou2020metaDataset,ullah2022metaAlbum}, while leaving sophisticated algorithmic and ensemble-based approaches for future researchers to evaluate on the benchmark.

\section{\dname{} Benchmark and Datasets}

\subsection{Motivation and Guiding Principles}\label{sec:motivation}
We first explain the motivating goals and guiding principles behind the design of \dname{}. The goal is to build a benchmark for multi-task meta-learning that will: (i) Encourage the community to develop meta-learners that are flexible enough to deal with greater task heterogeneity than before, and thus are more likely to be useful in practice with less curated episode distributions. This was identified as a major challenge in the discussion arising in several recent meta-learning and computer vision workshops and challenges\footnote{ICLR \url{https://sites.google.com/view/learning-2-learn/}, ECML \url{https://janvanrijn.github.io/metalearning/workshop2022}, NeurIPS \url{https://metalearning.chalearn.org/metadlneurips2021}, and ECCV  \url{http://www.ood-cv.org/}}. (ii) Ultimately progress on this benchmark should provide practical improvements in data-efficient learning for computer vision through the development of methods that can better transfer across different task types.

\begin{figure}[t]
  \centering
   \includegraphics[width=1.0\columnwidth]{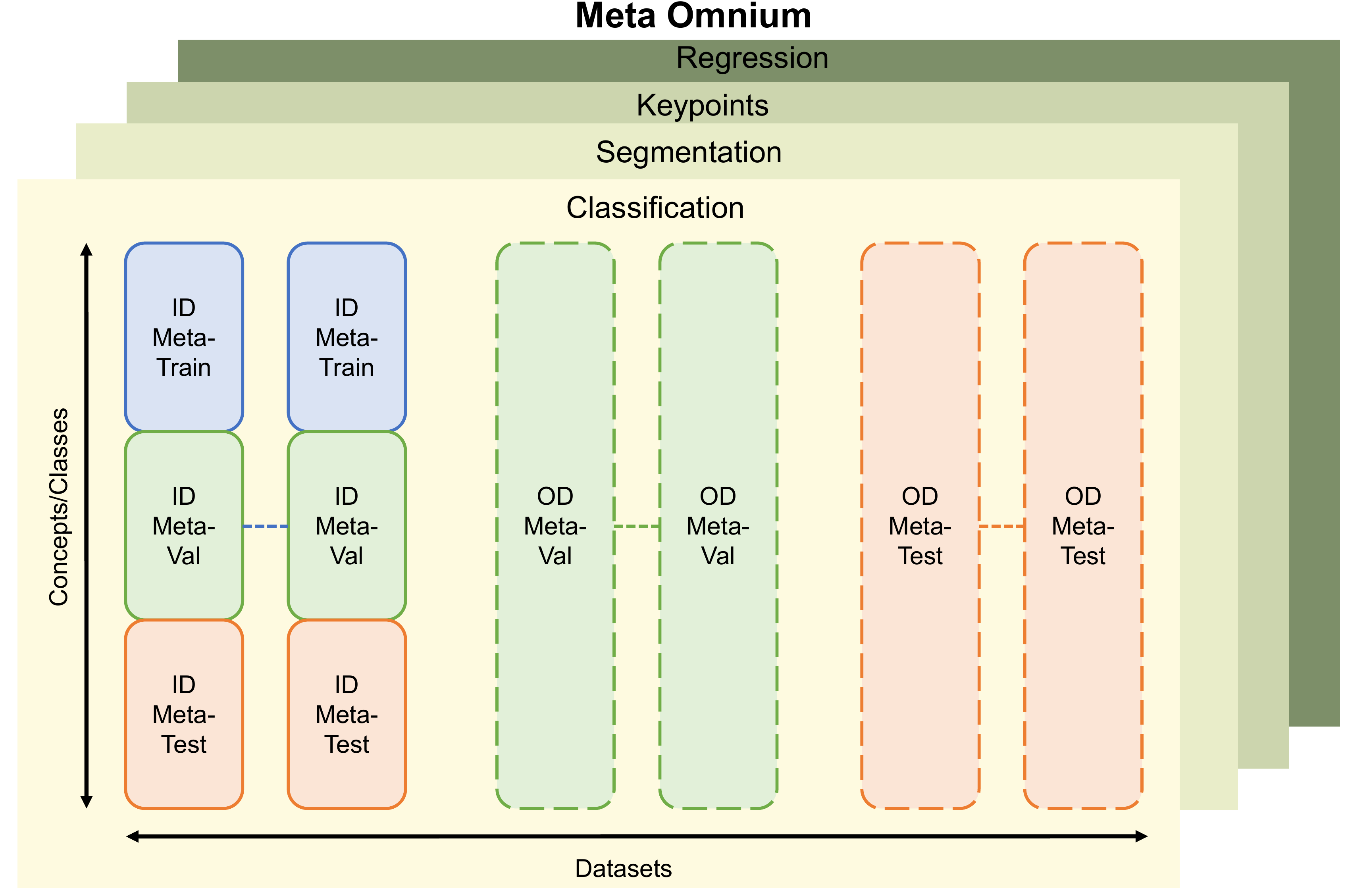}
   \caption{Schematic of benchmark and dataset splits. For each task, there are multiple datasets, which are divided into seen (solid border) and unseen (dash border) datasets. The seen (ID) datasets are divided class-wise into meta-train/meta-val/meta-test splits. The unseen datasets are held out for out-of-distribution (OD) evaluation. Meta-training is conducted on the ID-meta-train split of the seen datasets (blue). Models are validated on ID validation class splits, or OOD validation datasets (green). Results are reported on the ID test class splits and OOD test datasets (orange). We also hold out an entire task family (regression) for evaluating novel task generalisation.}
   \label{fig:schematic}
\end{figure}

In developing this benchmark, we established a few principles that we used to guide design choices. These include: (i) The benchmark should be lightweight in terms of storage and computing, making it accessible to a broad range of researchers and not only large corporations. (ii) The benchmark should cover multiple tasks with heterogeneous output spaces (as opposed to all classification, all regression, or all dense prediction), as well as multiple visual domains. In these regards, \dname{} is compared to alternatives in Table~\ref{tab:comparison}. (iii) The initial baselines should have only minimal \emph{task-specific decoders}. This is in contrast to the state of the art within various sub-disciplines of FSL such as segmentation \cite{min2021hypercorrelation, hong2022cost}, keypoint \cite{lu2022few, xiao2020few}, and classification \cite{afrasiyabi2022matching, ye2020few} where specially designed decoders are often used. This is to evaluate and encourage future research on \emph{learning-to-learn} across tasks, rather than primarily benchmarking how well we can manually engineer prior knowledge of optimal task-specific decoders. While we are not opposed to future competitors on this benchmark developing task-specific decoders, these should be evaluated separately against the minimal-decoder competitors. (iv) The benchmark should provide distinct datasets for in-distribution (ID) training and out-of-distribution (OOD) evaluation, to evaluate the robustness of the distribution-shift. This is already provided by \cite{triantafillou2020metaDataset,ullah2022metaAlbum} for classification, and we extend such an ID and OOD dataset ensemble to multiple tasks. Figure~\ref{fig:schematic} illustrates our dataset and task-split. (v) The benchmark should provide a clear hyper-parameter tuning protocol. With a number of recent studies showing that hyper-parameter tuning can dominate other effects of interest in computer vision \cite{gulrajani2021lostDG,musgrave2021unsupervisedDAreality,li2022findingDG}, this is important for a future-proof meta-learning benchmark. This is also related to the first cost point (i) above: Only for a benchmark with a modest cost can most institutions realistically expect to conduct hyper-parameter tuning. We provide the hyper-parameter tuning protocol. (vi) Finally, following the debate in \cite{tian2020rethinking,raghu2020rapidANIL,li2021metaCoding} as to the value of meta-learning vs conventional transfer learning, the dataset should support both episodic meta-learning and conventional transfer learning approaches. 

\subsection{Data Splits and Tasks} For each main task (classification, segmentation, keypoint localization), we split the datasets into seen datasets available for meta-training, and unseen datasets that are completely held out for out-of-domain meta-validation and meta-testing. Similarly to \cite{triantafillou2020metaDataset,ullah2022metaAlbum}, for the seen datasets, we construct category-wise splits into meta-train/val/test. While for the unseen datasets, there is no category-wise split as episodes from all categories from the whole dataset will be used for validation and testing respectively. The overall split organization is illustrated in Figure~\ref{fig:schematic}. We additionally have a completely held-out task: regression. Datasets from this task are not used during meta-training. 

Our multi-task setup enables us to define and compare two training protocols: \textbf{Single-task meta-learning} which evaluates how well meta-learning performs when trained and tested within a particular task family (within each plate in Figure~\ref{fig:schematic}); and \textbf{Multi-task meta-learning} which evaluates how well meta-learning performs when trained across all available task families (across all plates in Figure~\ref{fig:schematic}).

With this organization we can separately evaluate:
\textbf{Within-distribution generalization (ID)}: How well do meta-learners generalize to novel test concepts within the seen datasets?; and \textbf{Out-distribution generalization (OOD)}: How well do meta-learners generalize to novel concepts in unseen datasets? 

We provide two sources of validation data: ID and OOD, and our models are selected based on the combined performance across both. OOD validation is not supposed by the most popular Meta-Dataset benchmark  \cite{triantafillou2020metaDataset} as despite its larger size it does not provide OOD validation datasets.

\subsection{Datasets and Metrics} Given the considerations in Section~\ref{sec:motivation}, our benchmark consists of three main tasks (classification, segmentation, keypoints/pose) and one held-out task (regression). 

\noindent\textbf{Classification}\quad For classification we take the 10 datasets from the initial public release of Meta-Album \cite{ullah2022metaAlbum}. These images are all $128\times128$ and contain 19--706 classes per dataset, with 40 images per class. Three of these datasets are reserved for out-of-distribution meta-validation, and four for out-of-distribution meta-test.

\noindent\textbf{Segmentation}\quad For segmentation, we take FSS1000 \cite{li2020fss1000} for in-distribution (10,000 images, 1,000 classes), and combine it with VizWiz \cite{tseng2022vizwiz} for OOD meta-validation (862 images, 22 classes), and modified Pascal5i \cite{Shaban2017OneShotLF} (7,242 images, 6 classes) and the very distinct medical imaging dataset PH2 \cite{Mendona2013PH2A} (200 images, 3 classes) for meta-testing.  The segmentation images originally were of diverse sizes. We resize them all to $224\times224$ for \dname{}. Note that VizWiz and Pascal datasets originally contain more classes and images. We exclude the classes that overlap with that in the FSS1000 dataset for few-shot learning, and thus there are no classes overlapping among all the datasets.

\noindent\textbf{Keypoints}\quad For keypoints/pose, we take animal-pose \cite{Cao_2019_ICCV} for in-distribution, synthetic animal-pose \cite{Mu_2020_CVPR} for OOD meta-validation, and MPII human-pose \cite{andriluka14cvpr} for OOD meta-testing. All images are resized to $128\times128$. MPII includes about 40k people in over 25k images with annotated body keypoints. Animal Pose includes 5 animal categories for 6K instances in over 4k images. Each animal is cropped from the original image.
We keep cats and dogs for training, horses and sheep for in-domain validation, and cow for in-domain testing. Synthetic animal pose generates synthetic images using  animal CAD models rendered from various viewpoints and lightings on a random background. We keep only the horse and tiger categories in our final datasets.

\noindent\textbf{Regression}\quad For evaluating regression as a held-out task, we use four datasets corresponding to the test splits of \cite{gao2022metaRegression}: ShapeNet1D, ShapeNet2D, Distractor and Pascal1D \cite{yin2020metaMemorisation}. All images are resized to $128\times128$. ShapeNet1D aims to predict azimuth angles. It contains 30 categories in total and we keep the 3 categories from the test set. ShapeNet2D further includes 2D rotation with azimuth angles and elevation. The test set of ShapeNet2D contains 300 categories in total with 30 images per category. Distractor aims to predict the position of a target object in the presence of a distractor. It contains 12 categories in total and the test set has 2 categories. Each category contains 1000 objects with 36 images for each. Pascal1D aims to predict azimuth angle. The whole Pascal1D contains 65 objects from 10 categories. The test set contains 15 objects with 100 images for each object.
The supplementary material provides full details of all datasets and splits.

\subsection{Training API} For episodic learning, we proceed by (i) sampling a task, (ii) sampling a dataset, (iii) sampling an episode. Under our main protocol we consider variable 1 to 5-shot evaluation, but also evaluate separate 1 and 5 shot settings (training is always done with a variable number of shots -- support examples). For classification tasks, we follow  \cite{ullah2022metaAlbum} in generating 5-way episodes. For segmentation tasks, we follow  \cite{li2020fss1000} in considering each episode to be a binary foreground/background classification problem for a novel class and generate 2-way episodes. For keypoint, we form episodes by randomly selecting a class (e.g., animal category) and then randomly selecting a {subset of 5 keypoints} to localize for each episode. For regression tasks, we generate variable 5 to 25-shot episodes because it is a common practice to use more shots for regression tasks \cite{gao2022metaRegression}.

For non-episodic/transfer learning, we provide access to the meta-train portions of the seen datasets in conventional mini-batches for conventional single task and multi-task supervised learning.

\noindent\textbf{Evaluation Metrics}\quad For classification tasks, we use standard top-1 accuracy; for segmentation tasks, we use standard mean intersection-over-union (mIOU) that averages over IoU values of all object classes; for keypoint prediction, we report the Percentage of Correct Keypoints (PCK). In detail, a detected joint is considered correct if the distance between the predicted and the true joint is within a certain threshold. In our experiments, the threshold is 0.01 for normalized value, which stands for about 12.8 pixel of input image resolution. {For regression tasks, we follow \cite{gao2022metaRegression} and use the same metrics.}

\subsection{Architecture and Baseline Competitors} As discussed in Section~\ref{sec:motivation}, we aim to establish baselines that can be adapted to tasks with heterogeneous outputs, with minimal reliance on task-specific decoders. We follow \cite{triantafillou2020metaDataset,ullah2022metaAlbum} in using a ResNet-18  CNN \cite{he2016resNet} as a feature extractor architecture. For recognition tasks, we perform multi-class classification immediately after ResNet's GAP. For regression tasks, we perform linear regression directly after the ResNet's GAP. For keypoint tasks, we consider them to be a regression problem from the feature map to the keypoint location. For segmentation tasks, we use a simplified PSPNet-like \cite{zhao2017PSPNet} strategy. We concatenate the extracted feature maps from ResNet's feature pyramid, with upsampling where appropriate, to generate a feature map of size $w\times h$, and then do pixel-wise classification with $1 \times 1$ convolutional layer to obtain the final segmentation map. All tasks thus have only \emph{one} learnable weight as a minimal classifier/decoder after the common ResNet feature encoder. Based on this common encoder and minimal decoder architecture, we describe our meta-learning baselines. 
\noindent\textbf{Prototypical Network}\quad \cite{snell2017prototypical} is a classic meta-learner that exploits nearest-centroid metric learning for few-shot classification. ProtoNets were adapted to segmentation tasks in PANet \cite{wang2019panet}, by performing pixel-level feature matching between support prototypes and query pixels. We use the same principle together with the PSPNet-like features described earlier. To generalize ProtoNets to regression tasks such as keypoint prediction, we must relax the prototype assumption, and use them as simple Gaussian kernel-regression models \cite{bishop2006prml}. Specifically, we generate a feature embedding for each support example, and then for query examples, we calculate the negative exponential distance to the support examples, and use this inverse distance-weighted sum of support set labels as the prediction. Thus for regression tasks with a support set $S=\{(x_i,y_i)\}$ and query example $x_q$, ProtoRegression predicts
\begin{equation}
f(y_q|x_q,S)\propto\sum_i y_i\exp(-\tau(f_\theta(x_i)-f_\theta(x_q))^2)
\end{equation}
enabling us to learn deep feature $f_\theta$ in the usual episodic meta-learning way. We use cross-entropy loss for classification and segmentation, and MAE loss for regression tasks.

\noindent\textbf{DDRR}\quad Deep differentiable ridge-regression has been considered for few-shot recognition \cite{bertinetto2018closedFormMeta}, tracking \cite{zheng2020learningRRtracking}, and other tasks. It is related to ProtoNet in that the feature is not adapted after the meta-train stage, but different in that the decoder/classifier layer is learned by differentiable ridge-regression rather than nearest centroid or kernel regression. An elegant property of DDRR methods is that they naturally address regression tasks, although they have been repurposed for classification  \cite{bertinetto2018closedFormMeta} by conducting MSE-loss regression to a target 1-hot vector. Thus they are a natural choice for our benchmark. For application to segmentation, we apply DDRR in a $1\times1$ convolution-like way, to perform pixel classification for the output mask with a DDRR classifier at each pixel.
Further, we calibrate the prediction for binary cross entropy loss with a learnable scale and bias following \cite{bertinetto2018closedFormMeta}. DDRR uses MAE loss for only regression tasks and use MSE loss for other tasks. 

\noindent\textbf{MAML}\quad The seminal few-shot meta-learner MAML \cite{finn2017maml}  aims to learn an initial condition for per-episode gradient-descent. MAML is straightforward to adapt to different types of tasks. Based on each episode's support set, a new output layer is learned, and the feature extractor is updated, both by a few steps of gradient descent. Similarly to Meta-Dataset \cite{triantafillou2020metaDataset}, we do not learn an initialization for the output layer, since it can change size between episodes drawn from multiple tasks. To alleviate this challenge, we also follow Meta-Dataset in evaluating Proto-MAML -- a variant that initializes the MAML output layer based on the linear classifier/regressor suggested by nearest-centroid prior to gradient descent. Going beyond this, to adapt  Proto-MAML to regression tasks, we also initialize the output layer based on the ridge-regression solution to the support set. 

\noindent\textbf{Meta-Curvature}\quad Meta-Curvature \cite{park2019metaCurvature} is an enhancement of MAML that learns a pre-conditioning matrix to improve inner-loop adaptation, as well as an initial condition as in standard MAML. Meta-Curvature outperforms MAML in simpler single-task few-shot benchmarks. 

\noindent\textbf{Transfer Learning}\quad We also consider standard supervised learning on the meta-train tasks for transfer to the target tasks, a strategy reported to be competitive with meta-learning \cite{tian2020rethinking}. For adaptation, we explore both linear readout \cite{wang2019simpleshot,tian2020rethinking} and fine-tuning \cite{ullah2022metaAlbum}. Besides learning a new output layer from scratch, we also consider a fine-tuning version that initializes the classifier weights using class prototypes (recognition/segmentation) or ridge regression weights (keypoints/regression), inspired by Proto-MAML.

\noindent\textbf{Train-from-Scratch (TFS)}\quad We lastly consider training each episode from scratch using only the support set \cite{ullah2022metaAlbum}.

\subsection{Hyperparameter Optimization}
\label{sec:hpo}
As part of our benchmark, we perform hyperparameter optimization (HPO) to ensure we select appropriate hyperparameters for the diverse tasks that we consider. Multiobjective HPO under restricted resources is challenging, so we devise the following HPO protocol: estimate the performance of each candidate configuration on a lower fidelity (lower number of iterations) and then identify the configuration that works the best across all validation datasets considered (combination of in-domain and out-domain datasets, across various task families). Note that fast multi-fidelity methods such as Hyperband \cite{hyperband}, ASHA \cite{li2020asha} or PASHA \cite{bohdal2023pasha} are not applicable out of the box in our multi-objective setup, so we decided to train each candidate configuration using fixed 5,000 training iterations. Since different tasks/datasets are of different difficulties (and use different metrics), we normalize the score of each configuration for each validation dataset by the best score for that dataset across all candidate configurations. We then select the configuration with the best average normalized score.

Note that due to resource constraints we are only able to sample a relatively smaller number of candidates (30), so we utilize a sample efficient state-of-the-art Multi-Objective TPE method \cite{Ozaki2020MultiobjectiveProblems}, available from the Optuna library \cite{optuna_2019}. We perform HPO for multi-task and single-tasks setups separately, so single task classification, segmentation and keypoint estimation have their own set of hyperparameters; and the multi-task case has its own set. The hyperparameters include the meta-learning rate and optimizer, momentum, and various method-specific hyperparameters (full details are in the supplementary). Once the hyperparameters are chosen, we perform standard training of the model for the full number of iterations.

\section{Experiments}

\begin{table*}[t]
\centering
\begin{tabular}{clccccccccc}
\toprule
&    & \multicolumn{2}{c}{\textbf{Classification}}         & \multicolumn{2}{c}{\textbf{Segmentation}}             & \multicolumn{2}{c}{\textbf{Keypoints}}                                   & \multicolumn{3}{c}{\textbf{Average Rank}}           \\
 &              & ID                & OOD                      & ID                & OOD                       & ID                & OOD                                                & ID & OOD & AVG                       \\
               \midrule
\multirow{9}{*}{\rotatebox{90}{Single-Task}} & MAML & 58.7 & 61.6 & 54.7 & 42.1 & 25.4 & 33.0 & 4.3 & 3.3 & 3.8 \\
& Proto-MAML & 50.5 & 49.7 & 46.4 & 44.1 & 23.6 & 22.5 & 6.0 & 6.3 & 6.2 \\
& Meta-Curvature & 64.8 & 61.4 & 65.6 & 49.8 & 43.5 & 16.0 & 2.0 & 4.3 & 3.2 \\
& ProtoNet & 70.4 & 59.4 & 75.8 & 57.2 & 27.8 & 33.3 & \textbf{1.3} & \textbf{1.7} & \textbf{1.5} \\
& DDRR & 63.1 & 58.7 & 66.7 & 48.0 & 20.5 & 31.9 & 4.7 & 3.7 & 4.2 \\
\cline{2-11}
& Proto-FineTuning & 50.8 & 50.7 & 60.0 & 43.4 & 21.3 & 33.1 & 5.3 & 4.3 & 4.8 \\
& FineTuning & 42.3 & 48.2 & 50.5 & 40.0 & 25.7 & 30.0 & 5.7 & 6.7 & 6.2 \\
& Linear-Readout & 48.6 & 53.4 & 34.0 & 22.7 & 22.1 & 26.9 & 7.3 & 6.7 & 7.0 \\
& TFS & 31.5 & 42.0 & 42.8 & 37.6 & 21.0 & 26.0 & 8.3 & 8.0 & 8.2 \\
 \midrule
\multirow{9}{*}{\rotatebox{90}{Multi-Task}} & MAML & 59.1 & 58.5 & 43.3 & 37.4 & 24.3 & 23.9 & \textbf{2.7} & 4.7 & 3.7 \\
& Proto-MAML & 58.5 & 63.7 & 53.0 & 43.2 & 21.6 & 33.3 & 3.0 & \textbf{1.7} & \textbf{2.3} \\
& Meta-Curvature & 70.4 & 66.9 & 42.6 & 34.5 & 18.2 & 25.3 & 4.3 & 4.7 & 4.5 \\
& ProtoNet & 65.9 & 58.8 & 63.3 & 49.7 & 20.1 & 33.0 & \textbf{2.7} & 2.0 & \textbf{2.3} \\
& DDRR & 52.8 & 51.9 & 40.4 & 37.3 & 22.8 & 30.1 & 5.0 & 4.7 & 4.8 \\
\cline{2-11}
& Proto-FineTuning & 52.4 & 53.2 & 44.8 & 37.8 & 21.2 & 30.0 & 4.3 & 4.0 & 4.2 \\
& FineTuning & 44.1 & 51.2 & 41.3 & 36.1 & 18.1 & 20.5 & 7.7 & 7.0 & 7.3 \\
& Linear-Readout & 46.0 & 50.9 & 41.5 & 32.6 & 19.9 & 23.5 & 6.3 & 8.0 & 7.2 \\
& TFS & 21.9 & 23.8 & 38.7 & 35.8 & 14.1 & 11.0 & 9.0 & 8.3 & 8.7 \\
\bottomrule
\end{tabular}

\caption{Main Results. Results are presented as averages across the datasets within each task type and separately for in-distribution (ID) and out-of-distribution (OOD) datasets. Classification, segmentation, and keypoint results are reported in accuracy (\%), mIOU (\%), and PCK (\%) respectively. The upper and lower groups correspond to multi-task and single-task meta-training prior to evaluation on the same set of meta-testing episodes. Upper and lower sub row groups correspond to meta-learners and non-meta learners respectively. See the appendix for a full breakdown over individual datasets. }\label{tab:main}
\end{table*}
In this section, we aim to use our benchmark to answer the following questions: (1) \emph{Which meta-learner performs best on average across a heterogenous range of tasks?} Existing benchmarks have evaluated meta-learners for one task at a time, we now use our common evaluation platform to find out if any meta-learner can provide general-purpose learning to learn across different task types, or whether each task type prefers a different learner. Similarly, we can ask \emph{which meta-learner is most robust to out-of-distribution tasks?} (2) Having defined the first multi-task meta-learning benchmark, and generalizations of seminal meta-learners to different kinds of output spaces, we ask \emph{which meta-learner performs best for multi-task meta-learning?} More generally, \emph{is there a trend in gradient-based vs metric-based meta-learner success?} (3) \emph{Does multi-task meta-learning improve or worsen performance compared to single-task?} The former obviously provides more meta-training data, which should be advantageous, but the increased heterogeneity across meta-training episodes in the multi-task case also makes it harder to learn \cite{vuorio2019multimodalMAML,triantafillou2020metaDataset}. (4) \emph{How does meta-learning perform compared to simple transfer learning, or learning from scratch?}

\subsection{Experimental Settings}
We train each meta-learner for 30,000 meta-training iterations, with meta-validation after every 2,500 iterations (used for checkpoint selection). For evaluation during meta-testing we use 600 tasks for each corresponding dataset, and for meta-validation we use 1200 tasks together. We use random seeds to ensure that the same tasks are used across all methods that we compare. For transfer learning approaches (fine-tuning, training from scratch, etc.), we use 20 update steps during evaluation. We only retain the meta-learned shared feature extractor across tasks, and for each new evaluation task, we randomly initialize the output layer so that we can support any number of classes  as well as novel task families during meta-testing (in line with \cite{ullah2022metaAlbum}).

\subsection{Results} The main experimental results are shown in Table~\ref{tab:main}, where rows correspond to different few-shot learners, and columns report the average performance of test episodes, aggregated across multiple datasets in each task family, and broken down by "seen" datasets (ID) and "unseen" datasets (OOD). The table also reports the average rank of each meta-learner across each dataset, both overall and broken down by ID and OOD datasets. More specifically, for each setting (e.g. cls. ID) we calculate the rank of each method (separately for STL and MTL), and then we average
those ranks across cls., seg. and keypoints. From the results, we can draw the following main conclusions: 

\noindent(1) \emph{ProtoNet is the most versatile meta-learner}, as shown by its highest average rank in the single-task scenario. This validates our novel Kernel Regression extension of ProtoNet for tackling regression-type keypoint localization tasks. Somewhat surprisingly, \emph{ProtoNet is also the most robust to out-of-distribution episodes} (OOD) which is different from the conclusion of \cite{triantafillou2020metaDataset} and others who suggested that gradient-based adaptation is crucial to adapt to OOD data. However, it is also in line with the results of \cite{ullah2022metaAlbum} and the strong performance of prototypes more broadly \cite{bohdalFeedForwardPrototypes}. 

\noindent(2) Coming to multi-task meta-learning the situation is similar in that ProtoNet dominates the other competitors, but now sharing the first place with Proto-MAML.

\begin{figure*}[t]
  \centering
   \includegraphics[width=0.91\textwidth]{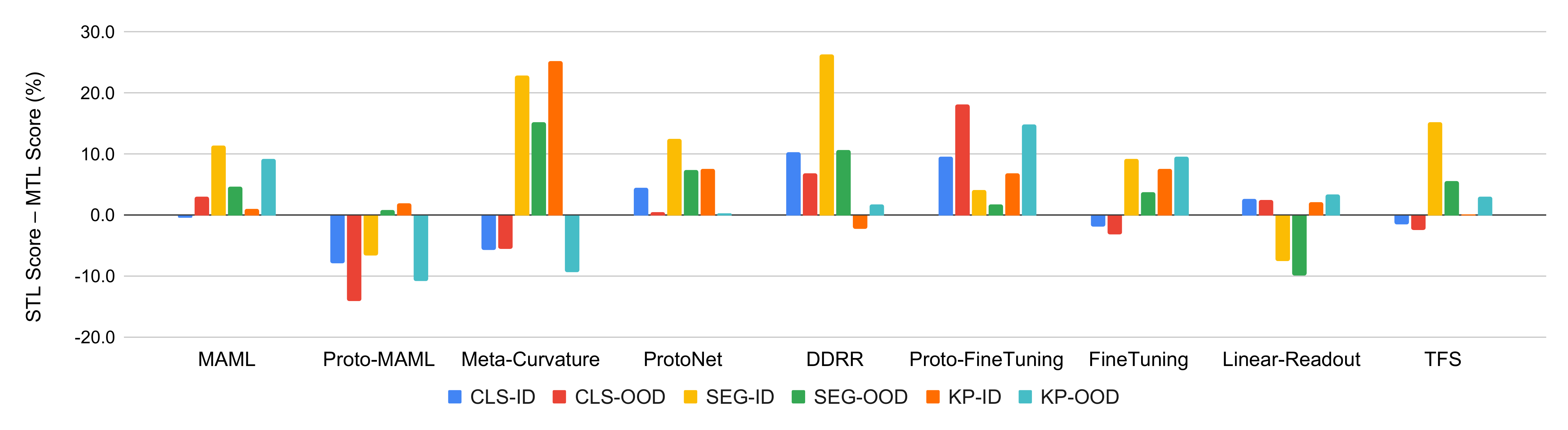}
   \caption{Analysis of the differences in scores between single-task (STL) and multi-task (MTL) learning for different methods.}
   \label{fig:stl_vs_mtl_analysis}
\end{figure*}

\noindent(3) To compare single-task and multi-task meta-learning (top and bottom blocks of Table~\ref{tab:main}) more easily, Figure~\ref{fig:stl_vs_mtl_analysis} shows the difference in meta-testing episode performance after STL and MTL meta-training for each method. Overall STL outperforms the MTL condition, showing that the difficulty of learning from heterogeneous tasks \cite{yu2020gradientSurgery,ruder2017mtlSurvey} outweighs the benefit of the extra available multi-task data.

\noindent(4) Finally, comparing meta-learning methods with simple transfer learning methods as discussed in \cite{tian2020rethinking,dumoulin2021a}, the \emph{best meta-learners are clearly better than transfer learning for both single and multi-task scenarios.}

We also note that Proto-MAML is better than MAML in the multi-task case, likely due to the importance of a good output-layer initialization in the case of heterogeneous episodes, as per \cite{triantafillou2020metaDataset}. Meta-Curvature outperforms MAML in single-task in-domain scenarios, in line with previous results \cite{park2019metaCurvature}, but it did not achieve stronger performance out-of-domain or in the multi-task case. Finally, while DDRR is perhaps the most elegant baseline in terms of most naturally spanning all task types, its overall performance is middling.

\subsection{Additional Analysis}

\noindent\textbf{How well can multi-task meta-learners generalize to completely new held-out tasks?}\quad We take the  multi-task meta-learners (trained on classification, segmentation, keypoints) and evaluate them on four regression benchmarks inspired by \cite{gao2022metaRegression}: ShapeNet1D, ShapeNet2D, Distractor and Pascal1D.
Because the metrics differ across datasets, we analyse the rankings and summarise the results in Table \ref{tab:ootreg}. We see the basic TFS performs the worst, with MAML, ProtoNets, DDRR being the best. However, in several cases the results were not better than predicting the mean (full results in the appendix), showing that learning-to-learn of completely new task families is an open challenge. 

\begin{table}[h]
\centering
 \begin{adjustbox}{max width=\linewidth}
\begin{tabular}{ccccccccc}
\hline
MAML & PMAML & MC & PN & DDRR & PFT & FT & LR & TFS \\
\hline
3.3 & 6.5 & 4.8 & 3.5 & 3.5 & 3.8 & 5.8 & 4.3 & 8.5 \\
\hline
\end{tabular}
\end{adjustbox}
\caption{Average ranking of the different methods across four out-of-task regression datasets.}\label{tab:ootreg}
\end{table}

\noindent\textbf{How much does external pre-training help?}\quad Our focus is on assessing the efficacy of meta-learning rather than representation transfer, but we also aim to support researchers investigating the impact of representation learning on external data prior to meta-learning  \cite{zhai2019largeVTAB,dumoulin2021a,hu202pmt}. We therefore specify evaluation conditions where external data outside our defined in-distribution meta-training set is allowed. 

We take two high-performing approaches in the multi-task scenario, Proto-MAML and ProtoNet, and we investigate to what extent external pre-training helps. We use the standard ImageNet1k pre-trained ResNet18, prior to conducting our meta-learning pipeline as usual. We use the same hyperparameters as selected earlier for these models to ensure consistent evaluation, and ensure that the differences in performance are not due to a better selection of hyperparameters. The results in Table \ref{tab:pretr} show that pretraining is not necessarily helpful in the considered multi-task setting, in contrast to purely recognition-focused evaluations \cite{zhai2019largeVTAB,dumoulin2021a,hu202pmt}, which were unambiguously positive about representation transfer from external data.

\begin{table}[t]
\centering
\resizebox{\columnwidth}{!}{
\begin{tabular}{lcccccccccc}
\hline
\multirow{2}{*}{Method} & \multirow{2}{*}{Pretrain} & \multicolumn{2}{c}{\textbf{Cls.}}         & \multicolumn{2}{c}{\textbf{Segm.}}             & \multicolumn{2}{c}{\textbf{Keyp.}} \\
&& ID & OOD & ID & OOD & ID & OOD \\
\hline
Proto-MAML &\xmark & 58.5 & 63.7 & 53.0 & 43.2 & 21.6 & 33.3 \\
ProtoNet & \xmark & 66.0 & 58.8 & 63.3 & 49.7 & 20.1 & 33.0 \\
Proto-MAML &\checkmark & 63.9 & 62.7 & 56.2 & 45.3 & 21.8 & 33.3 \\
ProtoNet &\checkmark & 63.5 & 58.6 & 62.0 & 49.0 & 20.1 & 33.1 \\
\hline
\end{tabular}}
\caption{Analysis of the impact of external data pretraining for selected meta-learners in  multi-task learning condition. The results show that ImageNet pretraining does not necessarily help improve performance. Cls., Segm., Keyp. represent classification, segmentation and keypoint respectively.}\label{tab:pretr}
\end{table}

\noindent\textbf{Analysis of Runtimes}\quad We analyze the times that the different meta-learning approaches spend on meta-training, meta-validation and meta-testing in the multitask learning case of our benchmark. The results in Table \ref{tab:times} show that all experiments are relatively lightweight, despite the ambitious goal of our benchmark to learn a meta-learner that can generalize across various task families. Most notably we observe that ProtoNets are the fastest approach, alongside being the best-performing one. 
Note that fine-tuning and training from scratch are expensive during the test time as they use backpropagation with a larger number of steps.

\noindent\textbf{Discussion and Future Work}\quad In future \dname{} can be used in a variety of ways beyond benchmarking multi-task meta-learning per-se. These including: studying the multi-task optimisation \cite{yu2020gradientSurgery} in meta-learning, studying HPO for meta-learning, developing validation strategies in meta-learning (using ID vs OD val sets \cite{li2022findingDG}), and studying the benefit of task-specific decoders and external data. 

\begin{table}[t]
\centering
 \begin{adjustbox}{max width=\columnwidth}
\begin{tabular}{lccccccccc}
\hline
Method & Train Time & Val Time & Test Time & Total Time \\
\hline
MAML & 1.8h & 1.9h & 0.9h & \hphantom{0}5.0h \\
Proto-MAML & 1.9h & 1.9h & 0.9h & \hphantom{0}5.1h \\
Meta-Curvature & 3.4h & 2.6h & 1.3h & \hphantom{0}7.6h \\
ProtoNet & 0.8h & 0.4h & 0.2h & \hphantom{0}1.8h \\
DDRR & 1.4h & 0.6h & 0.3h & \hphantom{0}2.7h \\
Proto-FineTuning & 1.7h & 4.5h & 2.3h & \hphantom{0}8.9h \\
FineTuning & 1.5h & 8.1h & 4.9h & 14.9h \\
Linear-Readout & 1.2h & 5.1h & 2.8h & \hphantom{0}9.6h \\
TFS & 0.0h & 0.8h & 6.2h & \hphantom{0}7.0h \\
\hline
\end{tabular}
\end{adjustbox}
\caption{Analysis of times needed by different algorithms in the multitask  setting (using one NVIDIA 1080 Ti GPU and 4 CPUs).}\label{tab:times}
\end{table}

\section{Conclusion} We have introduced \dname{}, the first multi-task few-shot meta-learning benchmark for computer vision. The benchmark is challenging in multiple highly topical ways including requiring learning on heterogeneous task distributions, evaluating generalization to out-of-distribution datasets, and uniquely challenging meta-learners to learn-to-learn and transfer knowledge across tasks with heterogeneous output spaces. \dname{} is nevertheless lightweight enough to be of broad interest and use for driving future research, and even to support future research in hyper-parameter optimization for meta-learning.  

\vspace{0.1cm}
\noindent\textbf{Acknowledgements}\quad This work was supported by the UKRI and EPSRC grant numbers EP/S000631/1, EP/L016427/1 and EP/S02431X/1; the MOD University Defence Research Collaboration (UDRC) in Signal Processing; the Royal Academy of Engineering under the Research Fellowship programme; Samsung AI Center, Cambridge; China grant No.2020KJ010802 and No.BYSYZHKC2021115.

{\small
\bibliographystyle{ieee_fullname}
\bibliography{egbib}
}

\appendix

\clearpage

\section{Full Dataset Details}\label{sec:fullDataset}
We describe the full details of our multi-task meta-dataset in Table \ref{tab:dataset_info} and provide further high-level details in this section.
\begin{itemize}
    \item Classification datasets: We reuse datasets selected in the initial release of Meta-Album \cite{ullah2022metaAlbum}. We split BCT (microscopy -- bacteria) \cite{10.1371/journal.pone.0184554}, BRD (large-animals -- birds) \cite{birds} and CRS (vehicles -- cars) \cite{KrauseStarkDengFei-Fei_3DRR2013} datasets into meta-training, in-domain meta-validation and in-domain meta-testing splits. We perform the splits randomly and in terms of classes -- 70\% for training, 15\% validation and testing each. FLW (plants -- flowers) \cite{Nilsback08}, MD-Mix (OCR) \cite{sun2021omniprint} and PLK (small animals -- plankton) \cite{whoiplankton} datasets are used for out-domain meta-validation. PLT-VIL (plant diseases) \cite{G2019323}, RESISC (remote sensing) \cite{DBLP:journals/corr/ChengHL17}, SPT (human actions -- sports) \cite{100-sports} and TEX (manufacturing -- textures) \cite{Fritz2004THEKD,Mallikarjuna2006THEK2,Kylberg2011c,lazebnik:inria-00548530} for out-domain meta-testing. We use the middle version (``Mini'') of these datasets as processed by the authors of Meta-Album \cite{ullah2022metaAlbum}, which allows us to keep the overall size of \dname sufficiently small.  
    \item Segmentation datasets: We first split FSS1000 \cite{li2020fss1000} dataset into in-domain train, validation, and test sets, i.e. FSS1000-Trn, FSS1000-Val, FSS1000-Test. We use Vizwiz \cite{tseng2022vizwiz} dataset for out-of-domain validation, and a modified version of Pascal 5i \cite{Shaban2017OneShotLF} and PH2 \cite{Mendona2013PH2A} datasets for out-of-domain testing. We exclude the object classes from the out-of-domain datasets that overlap with FSS1000 to ensure the classes during validation and testing are never seen during training.
    \item Keypoint estimation datasets: We use three keypoint datasets in the paper, including animal pose \cite{Cao_2019_ICCV}, synthetic animal pose \cite{Mu_2020_CVPR} and human pose \cite{andriluka14cvpr}. A single animal/human image is cropped from the original picture according to absolute maximum and minimum keypoint coordinates. The boundary is extended with 5 more pixels to avoid losing important information at object edges. Different keypoint datasets would have various target keypoints, so we cannot have a trivial solution like classification with a $N$-way $K$-shot setting, which stands for sampling $K$ samples from $N$ categories. Instead, we sample each keypoint task from one object category with only a fixed number of keypoints. In detail, we randomly select 5 keypoints per task, and train and fit the model to predict only 5 keypoints. This method leads to a general meta-learning keypoint prediction model that learns to predict corresponding keypoints from the limited support labels, which makes it possible for an arbitrary number of keypoint prediction tasks when conducted on more complex keypoint datasets.
    \item Regression datasets: We use regression datasets only for out-of-task (OOT) meta-test evaluation, so they are not used during meta-training. More specifically we use ShapeNet1D \cite{gao2022metaRegression}, ShapeNet2D \cite{gao2022metaRegression}, Distractor \cite{gao2022metaRegression} and Pascal1D datasets \cite{yin2020metaMemorisation}. Because regression problems typically require larger number of examples for adaptation, we use 5-times as many support examples compared to the other cases (e.g. instead of 5-shot we have 25-shot case). For our analysis experiments we consider the equivalent of variable 1-to-5-shot setting: variable 5-to-25-shot setting.
\end{itemize}

\begin{table*}[hbt!]
\centering
\resizebox{1.0\textwidth}{!}{
    \begin{tabular}{c|ccccccc}
 \toprule 
         Task Family & Dataset Name & Domain & \# Classes & \# Images & Cardinality & Role & Size (MB)  \\
         \midrule
         \multirow{16}{*}{\rotatebox{90}{Classification}} & BCT-Trn & Microscopy & 23 & 920 & (5) & Meta-train & 8\\
         & BRD-Trn & Bird & 220 & 8800 & (5) & Meta-train & 72\\
         & CRS-Trn & Car & 137 & 5480 & (5) & Meta-train & 44\\
         & BCT-Val & Microscopy & 5 & 200 & (5) & ID Meta-val & 1.7\\
         & BRD-Val & Bird & 47 & 1880 & (5) & ID Meta-val  & 15\\
        & CRS-Val & Car & 29 & 1160 & (5) & ID Meta-val  & 9\\
        & FLW & Flowers & 102 & 4080 & (5) & OD Meta-val  & 39\\
        & MD-MIX & OCR & 706 & 28240 & (5) & OD Meta-val  & 479\\
        & PLK & Plankton & 86 & 3440 & (5) & OD Meta-val  & 36\\
        & BCT-Test & Microscopy & 5 & 200 & (5) & ID Meta-test  & 1.7\\
        & BRD-Test & Bird & 48 & 1920 & (5) & ID Meta-test & 16\\
        & CRS-Test & Car & 30 & 1200 & (5) & ID Meta-test  & 10\\
        & PLT-VIL & Plant Disease & 38 & 1520 & (5) & OD Meta-test  & 14\\
        & RESISC & Remote Sensing & 45 & 1800 & (5) & OD Meta-test & 17\\
        & SPT & Sports & 73 & 2920 & (5) & OD Meta-test & 27\\
        & TEX & Textures & 64 & 2560 & (5) & OD Meta-test & 26\\
        \midrule
         \multirow{7}{*}{\rotatebox{90}{Segmentation}} & FSS1000-Trn & Natural Image & 520 & 5200 & (2) & Meta-train & 331\\
        & FSS1000-Val & Natural Image & 240 & 2400 & (2) & ID Meta-val & 150\\
        & FSS1000-Test & Natural Image& 240 & 2400 & (2) & ID Meta-test & 53\\
        & Pascal 5i & Natural Image & 6 & 7247 & (2) & OD Meta-test & 563\\
        & Vizwiz & Natural Image & 22 & 862 & (2) & OD Meta-val & 24\\
        & PH2 (Skin) & Medical Image & 3 & 200 & (2) & OD Meta-test & 114\\
        \midrule
         \multirow{6}{*}{\rotatebox{90}{\makecell{Keypoint\\Regression}}} & Animal pose - Trn & Animal & 2 & 3237 & (20, 2) & Meta-train & 112  \\
        & Animal pose - Val & Animal  & 2 & 2038 & (20, 2) & ID Meta-val  & 54\\
        & Animal pose - Test & Animal & 1 & 842 & (20, 2) & ID Meta-test & 18  \\
        & Synthetic Animal Pose &Synthetic Animal & 2 & 20000 & (18, 2) & OD Meta-val & 627 \\
        & MPII & Human & 1 & 28882 & (16, 2)  & OD Meta-test & 265\\
         \midrule
         \multirow{4}{*}{\rotatebox{90}{\makecell{Regression}}} & ShapeNet1D-Test & Synthetic Image & 60 & 3000 & (2) & OOT Meta-test & 8\\
        & ShapeNet2D-Test & Synthetic Image & 300 & 9000 & (4) & OOT Meta-test & 29\\
        & Distractor-Test & Synthetic Image& 200 & 7200 & (2) & OOT Meta-test & 93\\
        & Pascal1D-Test & Synthetic Image & 15 & 1500 & (1) & OOT Meta-test & 4\\

        \bottomrule
    \end{tabular}}
    \caption{Details of all task families included in \dname. }
    \label{tab:dataset_info}
\end{table*}

\section{Additional Analysis}

\begin{figure}[t]
    \centering
    \includegraphics[width=\linewidth]{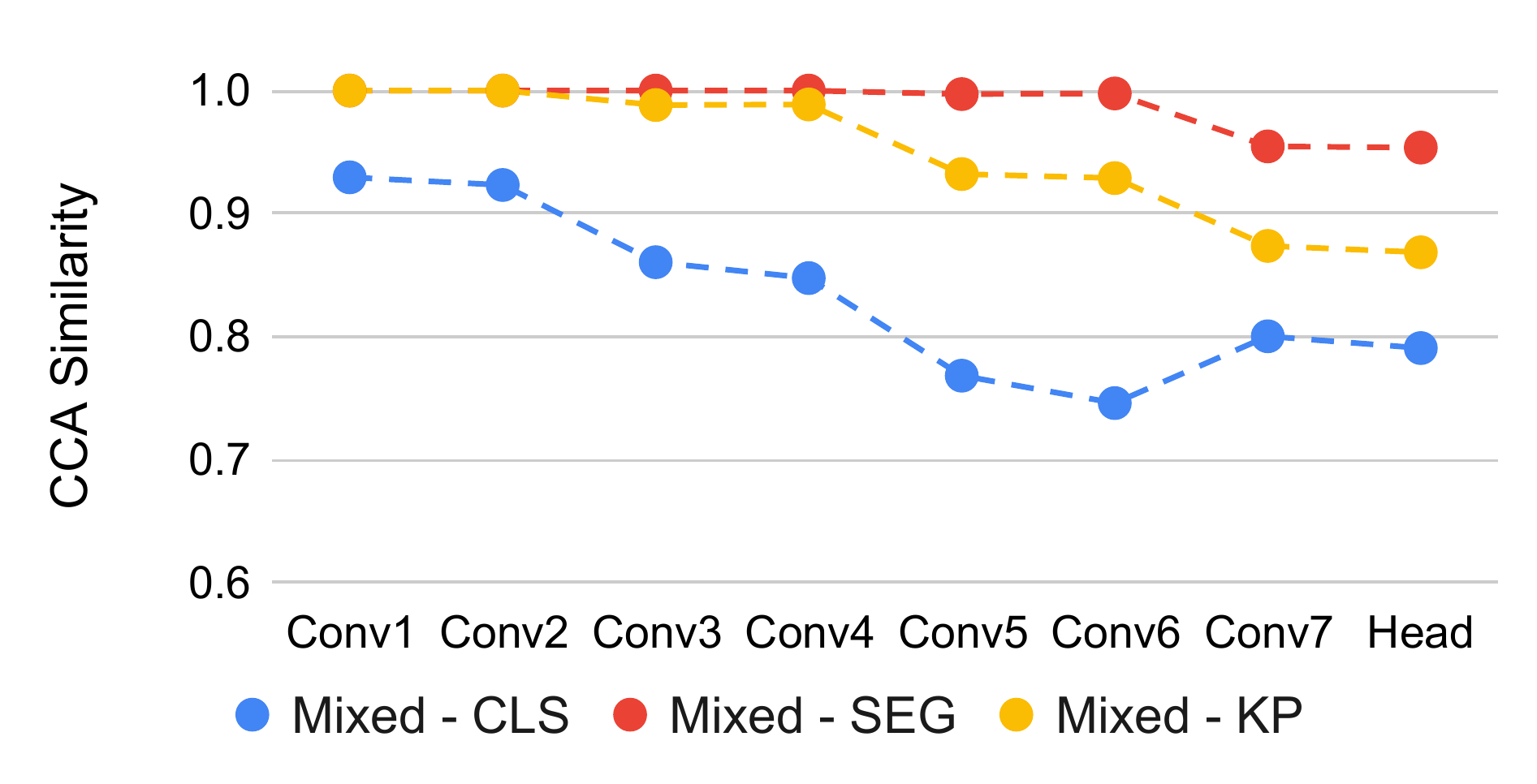}
    \caption{Analysis of the layer adaptation by MAML in \dname{}.}
    \label{fig:layerAdapt}
\end{figure}

\noindent\textbf{How do gradient-based meta-learners adapt their layers?}\quad A recent debate in few-shot meta-learning has been around whether gradient-based meta-learners really learn to adapt, or simply reuse features without adaptation. \cite{raghu2020rapidANIL}  claimed that feature reuse was the dominant effect after measuring the representational change pre- and post-adaptation and finding that representational change was primarily in the output layer. We analyze this using Canonical Correlation Analysis (CCA) \cite{raghu2017svcca,morcos2018insights} for \dname{}, reporting the representational change of multi-task MAML by layer for each task family during meta-testing. From the results in Figure~\ref{fig:layerAdapt}, we observe that: (1) The degree of representational change varies substantially with tasks, (2) Similar to \cite{raghu2020rapidANIL}, there is greater representational change at the later layers, especially the final output layer. However, significant amount of adaptation is done also in the earlier layers, which we attribute to the greater diversity of tasks and visual domains in \dname{} compared to the simple recognition episodes in miniImageNet studied by \cite{raghu2020rapidANIL}.

\section{Additional Experimental Details}

\subsection{Hyperparameter Optimization (HPO)}
The details of how we perform HPO are described in Section \ref{sec:hpo}, and in this section we provide additional details. The search space for HPO is as follows (note that momentum is only used if SGD optimizer is selected):
\begin{itemize}
    \item MAML and Meta-Curvature: meta-learning rate $\in (10^{-4}, 10^{-1})$ (log scale), meta optimizer $\in \{\text{Adam}, \text{SGD}\}$, momentum $\in \{0.0, 0.9, 0.99\}$, inner-loop learning rate $\in (10^{-3}, 0.5)$ (log scale)
    \item Proto-MAML: same as MAML and also parameter $\lambda \in (0.01, 100)$ (log scale) that influences the prototype calculation in the case of keypoint estimation
    \item ProtoNet: meta-learning rate $\in (10^{-4}, 10^{-1})$ (log scale), meta optimizer $\in \{\text{Adam}, \text{SGD}\}$, momentum $\in \{0.0, 0.9, 0.99\}$ and distance temperature $\in (0.1, 10.0)$ (log scale) that is used for keypoint estimation
    \item DDRR: meta-learning rate $\in (10^{-4}, 10^{-1})$ (log scale), meta optimizer $\in \{\text{Adam}, \text{SGD}\}$, momentum $\in \{0.0, 0.9, 0.99\}$ and $\lambda \in (0.01, 100)$ (log scale)
    \item Proto-FineTuning: learning rate $\in (10^{-4}, 10^{-1})$ (log scale), optimizer $\in \{\text{Adam}, \text{SGD}\}$, momentum $\in \{0.0, 0.9, 0.99\}$ and $\lambda \in (0.01, 100)$ (log scale)
    \item FineTuning: learning rate $\in (10^{-4}, 10^{-1})$ (log scale), optimizer $\in \{\text{Adam}, \text{SGD}\}$ and momentum $\in \{0.0, 0.9, 0.99\}$
    \item Linear-Readout and TFS: same as FineTuning
\end{itemize}

After training a model with the candidate configuration for 5,000 iterations, we evaluate its validation performance. We use 100 tasks for evaluating the in-domain validation performance, and additional 100 tasks for evaluation of out-domain performance. As part of our multi-objective HPO, we minimize the validation error rates (or appropriate equivalent) and use each dataset as a separate objective. We perform HPO on the primary variable 1-to-5 shot setting. We use the same hyperparameters also for the 1-shot and 5-shot settings.

Our HPO is reasonably fast, and it generally takes between a few hours up to two days in the slowest cases (using a single NVIDIA 1080 Ti GPU with 12GB memory and using 4 CPUs). As a result, it is feasible to run the HPO even with modest resources when designing new approaches for our multi-task scenario. We provide the found hyperparameters within the released code.

\subsection{Experimental Settings}
Many of our experimental settings follow Meta-Album \cite{ullah2022metaAlbum}), whose code-base we have also used as the starting point. All approaches use one task in a meta-batch. We use 5 inner-loop steps during meta-training and 10 inner-loop steps during evaluation for MAML, Proto-MAML and Meta-Curvature. We use gradient-clipping of 5. DDRR uses an adjustment layer, the scale of which is initialized to 5.0 (with the adjust base set to 1.0). Proto-FineTuning, FineTuning, Linear-Readout and TFS use 20 fine-tuning steps during evaluation. The training minibatch size for these approaches is 16, while the testing minibatch size is 4. We use standard ImageNet normalization for segmentation tasks, but we do not use normalization in the other cases, following earlier work \cite{ullah2022metaAlbum}.

We train each model for 30,000 iterations and evaluate the model on validation data after every 2,500 tasks, including at the beginning and the end (used for early stopping -- model selection). We use 5-way tasks during both training and evaluation. The number of shots is between 1 and 5 during meta-training, and we consider 3 setups for evaluation: variable 1-to-5-shot (primary), 1-shot and 5-shot (presented in the appendix). The query size is 5 examples per category and this has been selected to be consistent across the different datasets. Validation uses 600 tasks for each of in-domain and out-domain evaluation. Testing uses 600 tasks per dataset to provide more rigorous evaluation. 

During evaluation, we randomly initialize the top layer weights (classifier) to enable any-way predictions, in line with previous literature \cite{ullah2022metaAlbum}. We do this for the approaches that perform fine-tuning (e.g. MAML or Fine-Tuning baseline). Note that in approaches such as Proto-MAML the top layer is initialized using weights derived from the prototypes or ridge regression solution.

\section{Detailed Per-Dataset Results}
We include detailed per-dataset results (various-shot evaluation), first showing the single-task learning results for classification, segmentation and keypoint estimation, followed by multi-task learning results. In each case, we separately report the results for in-domain and out-of-domain evaluation. We also include detailed results for our out-of-task evaluation using regression datasets. Summary 1-shot and 5-shot results are included for the single and multi-task settings.

\section{Full Acknowledgements}
This work was supported by the Engineering and Physical Sciences Research Council (EPSRC) grant number EP/S000631/1; the MOD University Defence Research Collaboration (UDRC) in Signal Processing; EPSRC Centre for Doctoral Training in Data Science, funded by the UK Engineering and Physical Sciences Research Council (grant EP/L016427/1) and the University of Edinburgh; the United Kingdom Research and Innovation (grant EP/S02431X/1), UKRI Centre for Doctoral Training in Biomedical AI at the University of Edinburgh, School of Informatics, and Samsung AI Center, Cambridge. This project was supported by the Royal Academy of Engineering under the Research Fellowship programme. This work was also supported by Key Project Plan of Blockchain in Ministry of Education of the People's Republic of China (Grant No. 2020KJ010802), Innovation and Transformation Fund of Peking University Third Hospital (Grant No. BYSYZHKC2021115) and China Scholarship Council. Yongshuo Zong is supported by the United Kingdom Research and Innovation (grant EP/S02431X/1), UKRI Centre for Doctoral Training in Biomedical AI at the University of Edinburgh, School of Informatics. For the purpose of open access, the author has applied a creative commons attribution (CC BY) licence to any author accepted manuscript version arising.

\begin{table*}[t]
\centering
\begin{tabular}{lccc}
\hline
Method & BCT-Test & BRD-Test & CRS-Test \\
\hline
MAML & 78.09 $\pm$ 0.75 & 64.14 $\pm$ 1.17 & 33.87 $\pm$ 0.94 \\
Proto-MAML & 74.29 $\pm$ 0.83 & 51.99 $\pm$ 1.16 & 25.25 $\pm$ 0.61 \\
Meta-Curvature & 85.31 $\pm$ 0.66 & 71.99 $\pm$ 1.09 & 37.19 $\pm$ 0.95 \\
ProtoNet & 81.53 $\pm$ 0.66 & 75.39 $\pm$ 1.07 & 54.35 $\pm$ 1.12 \\
DDRR & 76.49 $\pm$ 0.80 & 69.25 $\pm$ 1.15 & 43.52 $\pm$ 1.03 \\
\hline
Proto-FineTuning & 42.92 $\pm$ 0.89 & 68.69 $\pm$ 1.17 & 40.81 $\pm$ 1.03 \\
FineTuning & 41.48 $\pm$ 0.85 & 52.51 $\pm$ 1.15 & 32.97 $\pm$ 0.78 \\
Linear-Readout & 45.44 $\pm$ 0.87 & 64.33 $\pm$ 1.12 & 36.11 $\pm$ 0.80 \\
TFS & 34.19 $\pm$ 0.86 & 35.14 $\pm$ 0.93 & 25.25 $\pm$ 0.71 \\
\hline
\end{tabular}
\caption{In-domain single-task classification results. Mean test accuracy (\%) and 95\% confidence interval across test tasks.}\label{tab:detailed_results}
\end{table*}

\begin{table*}[t]
\centering
\begin{tabular}{lcccc}
\hline
Method & PLT\_VIL & RESISC & SPT & TEX \\
\hline
MAML & 62.69 $\pm$ 1.14 & 51.83 $\pm$ 1.06 & 46.24 $\pm$ 1.05 & 85.49 $\pm$ 1.02 \\
Proto-MAML & 46.59 $\pm$ 1.00 & 39.79 $\pm$ 0.94 & 35.24 $\pm$ 0.91 & 77.11 $\pm$ 1.14 \\
Meta-Curvature & 61.88 $\pm$ 1.07 & 52.00 $\pm$ 1.13 & 45.11 $\pm$ 1.06 & 86.55 $\pm$ 0.98 \\
ProtoNet & 59.68 $\pm$ 1.15 & 51.17 $\pm$ 1.04 & 43.65 $\pm$ 1.06 & 83.02 $\pm$ 1.03 \\
DDRR & 60.28 $\pm$ 1.19 & 48.70 $\pm$ 1.04 & 42.83 $\pm$ 1.04 & 83.17 $\pm$ 1.10 \\
\hline
Proto-FineTuning & 51.50 $\pm$ 1.27 & 41.92 $\pm$ 1.06 & 39.54 $\pm$ 1.03 & 69.77 $\pm$ 1.39 \\
FineTuning & 46.83 $\pm$ 1.13 & 41.37 $\pm$ 0.96 & 36.03 $\pm$ 0.90 & 68.39 $\pm$ 1.23 \\
Linear-Readout & 52.68 $\pm$ 1.02 & 46.24 $\pm$ 1.00 & 41.39 $\pm$ 0.94 & 73.45 $\pm$ 1.13 \\
TFS & 43.87 $\pm$ 1.03 & 36.36 $\pm$ 0.90 & 35.07 $\pm$ 0.91 & 52.54 $\pm$ 1.33 \\
\hline
\end{tabular}
\caption{Out-of-domain single-task classification results. Mean test accuracy (\%) and 95\% confidence interval across test tasks.}\label{tab:detailed_results}
\end{table*}

\begin{table*}[t]
\centering
\begin{tabular}{lc}
\hline
Method & FSS1000-Test \\
\hline
MAML & 54.70 $\pm$ 1.68 \\
Proto-MAML & 46.40 $\pm$ 1.62 \\
Meta-Curvature & 65.57 $\pm$ 1.21 \\
ProtoNet & 75.84 $\pm$ 0.98 \\
DDRR & 66.71 $\pm$ 1.20 \\
\hline
Proto-FineTuning & 59.96 $\pm$ 1.55 \\
FineTuning & 50.52 $\pm$ 1.59 \\
Linear-Readout & 34.00 $\pm$ 1.85 \\
TFS & 42.80 $\pm$ 1.52 \\
\hline
\end{tabular}
\caption{In-domain single-task segmentation results. Mean test mIoU (\%) and 95\% confidence interval across test tasks. Larger mIoU is better.}\label{tab:detailed_results}
\end{table*}

\begin{table*}[t]
\centering
\begin{tabular}{lcc}
\hline
Method & Pascal 5i & PH2 \\
\hline
MAML & 15.27 $\pm$ 1.29 & 68.88 $\pm$ 1.25 \\
Proto-MAML & 22.80 $\pm$ 1.19 & 65.46 $\pm$ 1.19 \\
Meta-Curvature & 27.66 $\pm$ 1.22 & 71.92 $\pm$ 0.80 \\
ProtoNet & 36.49 $\pm$ 1.39 & 77.82 $\pm$ 0.79 \\
DDRR & 29.07 $\pm$ 1.12 & 66.95 $\pm$ 0.77 \\
\hline
Proto-FineTuning & 21.03 $\pm$ 1.24 & 65.79 $\pm$ 1.21 \\
FineTuning & 16.23 $\pm$ 1.24 & 63.68 $\pm$ 1.11 \\
Linear-Readout & \phantom{0}5.67 $\pm$ 0.80 & 39.67 $\pm$ 1.91 \\
TFS & 15.45 $\pm$ 1.03 & 59.77 $\pm$ 1.18 \\
\hline
\end{tabular}
\caption{Out-of-domain single-task segmentation results. Mean test mIoU (\%) and 95\% confidence interval across test tasks. Larger mIoU is better.}\label{tab:detailed_results}
\end{table*}

\begin{table*}[t]
\centering
\begin{tabular}{lc}
\hline
Method & Animal pose - Test \\
\hline
MAML & 25.36 $\pm$ 0.93 \\
Proto-MAML & 23.63 $\pm$ 0.84 \\
Meta-Curvature & 43.47 $\pm$ 0.99 \\
ProtoNet & 27.79 $\pm$ 0.89 \\
DDRR & 20.53 $\pm$ 0.72 \\
\hline
Proto-FineTuning & 21.27 $\pm$ 0.74 \\
FineTuning & 25.69 $\pm$ 0.90 \\
Linear-Readout & 22.09 $\pm$ 0.74 \\
TFS & 20.98 $\pm$ 0.63 \\
\hline
\end{tabular}
\caption{In-domain single-task keypoint estimation results. Mean test PCK (\%) and 95\% confidence interval across test tasks. Larger PCK is better.}\label{tab:detailed_results}
\end{table*}

\begin{table*}[t]
\centering
\begin{tabular}{lc}
\hline
Method & MPII \\
\hline
MAML & 33.04 $\pm$ 0.64 \\
Proto-MAML & 22.48 $\pm$ 0.64 \\
Meta-Curvature & 16.00 $\pm$ 0.39 \\
ProtoNet & 33.33 $\pm$ 0.71 \\
DDRR & 31.88 $\pm$ 0.63 \\
\hline
Proto-FineTuning & 33.10 $\pm$ 0.71 \\
FineTuning & 30.03 $\pm$ 0.53 \\
Linear-Readout & 26.86 $\pm$ 0.46 \\
TFS & 25.95 $\pm$ 0.52 \\
\hline
\end{tabular}
\caption{Out-of-domain single-task keypoint estimation results. Mean test PCK (\%) and 95\% confidence interval across test tasks. Larger PCK is better.}\label{tab:detailed_results}
\end{table*}

\begin{table*}[t]
\centering
\begin{tabular}{lccccc}
\hline
Method & FSS1000-Test & BCT-Test & BRD-Test & CRS-Test & Animal pose - Test \\
\hline
MAML & 43.31 $\pm$ 1.60 & 89.05 $\pm$ 0.61 & 59.94 $\pm$ 0.99 & 28.19 $\pm$ 0.75 & 24.25 $\pm$ 0.79 \\
Proto-MAML & 53.03 $\pm$ 1.51 & 84.71 $\pm$ 0.70 & 59.79 $\pm$ 1.12 & 30.87 $\pm$ 0.87 & 21.63 $\pm$ 0.76 \\
Meta-Curvature & 42.60 $\pm$ 1.74 & 85.43 $\pm$ 0.66 & 76.85 $\pm$ 1.06 & 48.97 $\pm$ 1.09 & 18.21 $\pm$ 0.47 \\
ProtoNet & 63.32 $\pm$ 1.09 & 81.95 $\pm$ 0.68 & 72.31 $\pm$ 1.05 & 43.58 $\pm$ 1.04 & 20.10 $\pm$ 0.75 \\
DDRR & 40.39 $\pm$ 1.11 & 77.19 $\pm$ 0.73 & 51.47 $\pm$ 1.11 & 29.59 $\pm$ 0.83 & 22.77 $\pm$ 0.73 \\
\hline
Proto-FineTuning & 44.80 $\pm$ 1.62 & 41.11 $\pm$ 0.91 & 71.21 $\pm$ 1.21 & 44.85 $\pm$ 1.06 & 21.16 $\pm$ 0.74 \\
FineTuning & 41.31 $\pm$ 1.74 & 42.84 $\pm$ 0.89 & 55.41 $\pm$ 1.22 & 34.05 $\pm$ 0.81 & 18.05 $\pm$ 0.49 \\
Linear-Readout & 41.53 $\pm$ 1.66 & 39.07 $\pm$ 0.78 & 63.60 $\pm$ 1.04 & 35.18 $\pm$ 0.81 & 19.89 $\pm$ 0.52 \\
TFS & 38.66 $\pm$ 1.56 & 22.45 $\pm$ 0.54 & 22.74 $\pm$ 0.49 & 20.39 $\pm$ 0.39 & 14.09 $\pm$ 0.75 \\
\hline
\end{tabular}
\caption{In-domain multi-task learning results. Mean test score (\%) and 95\% confidence interval across test tasks. Larger score is better in all cases.}\label{tab:detailed_results}
\end{table*}

\begin{table*}[t]
\centering
\resizebox{1.0\textwidth}{!}{
\begin{tabular}{lccccccc}
\hline
Method & PLT\_VIL & RESISC & SPT & TEX & Pascal 5i & PH2 & MPII \\
\hline
MAML & 60.81 $\pm$ 1.11 & 48.19 $\pm$ 1.04 & 39.23 $\pm$ 0.91 & 85.59 $\pm$ 1.02 & 15.57 $\pm$ 1.11 & 59.28 $\pm$ 1.32 & 23.85 $\pm$ 0.47 \\
Proto-MAML & 65.18 $\pm$ 1.18 & 54.04 $\pm$ 1.10 & 49.84 $\pm$ 1.09 & 85.85 $\pm$ 0.95 & 21.90 $\pm$ 1.16 & 64.51 $\pm$ 1.04 & 33.34 $\pm$ 0.68 \\
Meta-Curvature & 70.98 $\pm$ 1.09 & 56.05 $\pm$ 1.15 & 51.09 $\pm$ 1.18 & 89.63 $\pm$ 0.80 & 13.29 $\pm$ 1.12 & 55.78 $\pm$ 1.52 & 25.29 $\pm$ 0.39 \\
ProtoNet & 60.55 $\pm$ 1.10 & 50.13 $\pm$ 1.04 & 41.92 $\pm$ 1.05 & 82.71 $\pm$ 1.00 & 30.46 $\pm$ 1.11 & 68.95 $\pm$ 0.87 & 33.00 $\pm$ 0.69 \\
DDRR & 53.19 $\pm$ 1.13 & 41.49 $\pm$ 0.98 & 35.73 $\pm$ 0.98 & 77.05 $\pm$ 1.19 & 20.19 $\pm$ 0.68 & 54.35 $\pm$ 0.78 & 30.08 $\pm$ 0.59 \\
\hline
Proto-FineTuning & 55.05 $\pm$ 1.21 & 44.17 $\pm$ 1.05 & 41.79 $\pm$ 1.04 & 71.64 $\pm$ 1.33 & 15.19 $\pm$ 1.06 & 60.47 $\pm$ 1.30 & 30.04 $\pm$ 0.60 \\
FineTuning & 50.88 $\pm$ 1.16 & 43.59 $\pm$ 1.00 & 38.07 $\pm$ 0.94 & 72.41 $\pm$ 1.18 & 11.68 $\pm$ 1.02 & 60.58 $\pm$ 1.39 & 20.46 $\pm$ 0.33 \\
Linear-Readout & 49.78 $\pm$ 1.00 & 44.57 $\pm$ 0.98 & 40.83 $\pm$ 0.97 & 68.58 $\pm$ 1.12 & 14.37 $\pm$ 1.07 & 50.73 $\pm$ 1.96 & 23.47 $\pm$ 0.35 \\
TFS & 23.15 $\pm$ 0.51 & 23.01 $\pm$ 0.47 & 22.45 $\pm$ 0.49 & 26.63 $\pm$ 0.71 & 13.12 $\pm$ 0.96 & 58.41 $\pm$ 1.25 & 11.04 $\pm$ 0.29 \\
\hline
\end{tabular}}
\caption{Out-of-domain multi-task learning results. Mean test score (\%) and 95\% confidence interval across test tasks. Larger score is better in all cases.}\label{tab:detailed_results}
\end{table*}

\begin{table*}[t]
\centering
\begin{tabular}{lcccc}
\hline
Method & ShapeNet2D-Test & Distractor-Test & ShapeNet1D-Test & Pascal1D-Test \\
\hline
MAML & \phantom{0}95.44 $\pm$ 2.82 & 38.68 $\pm$ 0.60 & 54.20 $\pm$ 2.10 & \phantom{0}2.88 $\pm$ 0.10 \\
Proto-MAML & \phantom{0}69.94 $\pm$ 1.50 & 39.58 $\pm$ 0.69 & 63.17 $\pm$ 2.33 & 51.74 $\pm$ 1.15 \\
Meta-Curvature & \phantom{0}70.55 $\pm$ 2.32 & 38.73 $\pm$ 0.66 & 43.17 $\pm$ 2.03 & 12.04 $\pm$ 0.61 \\
ProtoNet & \phantom{0}63.50 $\pm$ 1.15 & 38.53 $\pm$ 0.58 & 84.53 $\pm$ 1.92 & \phantom{0}2.53 $\pm$ 0.06 \\
DDRR & \phantom{0}64.41 $\pm$ 1.64 & 41.67 $\pm$ 0.68 & 46.08 $\pm$ 1.90 & \phantom{0}2.11 $\pm$ 0.07 \\
\hline
Proto-FineTuning & \phantom{0}61.94 $\pm$ 2.25 & 39.44 $\pm$ 0.69 & 58.33 $\pm$ 2.54 & \phantom{0}4.17 $\pm$ 0.18 \\
FineTuning & \phantom{0}63.36 $\pm$ 1.54 & 51.52 $\pm$ 1.44 & 81.36 $\pm$ 2.12 & \phantom{0}6.54 $\pm$ 0.31 \\
Linear-Readout & \phantom{0}68.66 $\pm$ 1.94 & 40.53 $\pm$ 0.71 & 46.93 $\pm$ 2.05 & \phantom{0}2.62 $\pm$ 0.09 \\
TFS & 133.67 $\pm$ 2.67 & 95.36 $\pm$ 0.91 & 88.25 $\pm$ 1.88 & \phantom{0}6.63 $\pm$ 0.31 \\
\hline
\end{tabular}
\caption{Evaluation of multi-task models on out-of-task regression datasets, using variable 5-to-25-shot episodes. Lower value is better.}\label{tab:regressiondetailed}
\end{table*}

\begin{table*}[t]
\centering
\begin{tabular}{clccccccccc}
\toprule
&    & \multicolumn{2}{c}{\textbf{Classification}}         & \multicolumn{2}{c}{\textbf{Segmentation}}             & \multicolumn{2}{c}{\textbf{Keypoints}}                                   & \multicolumn{3}{c}{\textbf{Average Rank}}           \\
 &              & ID                & OOD                      & ID                & OOD                       & ID                & OOD                                                & ID & OOD & AVG                       \\
               \midrule
\multirow{9}{*}{\rotatebox{90}{Single-Task}} & MAML & 50.8 & 50.8 & 44.1 & 33.6 & 34.7 & 33.4 & 4.3 & 4.0 & 4.2 \\
& Proto-MAML & 53.5 & 52.4 & 46.0 & 39.2 & 23.5 & 14.8 & 4.7 & 4.7 & 4.7 \\
& Meta-Curvature & 58.0 & 51.6 & 60.5 & 40.1 & 38.1 & 16.1 & \textbf{1.7} & 4.3 & \textbf{3.0} \\
& ProtoNet & 61.7 & 50.2 & 73.7 & 52.5 & 22.5 & 31.9 & 2.7 & \textbf{3.7} & 3.2 \\
& DDRR & 54.7 & 48.9 & 60.1 & 42.2 & 22.1 & 32.2 & 4.7 & 4.0 & 4.3 \\
 \cline{2-11}
& Proto-FineTuning & 47.0 & 50.4 & 50.5 & 36.6 & 22.4 & 32.8 & 5.7 & 4.3 & 5.0 \\
& FineTuning & 35.5 & 43.2 & 42.4 & 36.6 & 34.6 & 33.8 & 6.0 & 4.7 & 5.3 \\
& Linear-Readout & 46.2 & 48.0 & 30.3 & 18.3 & 26.5 & 26.7 & 6.7 & 7.3 & 7.0 \\
& TFS & 27.2 & 36.4 & 31.5 & 30.3 & 19.5 & 20.0 & 8.7 & 8.0 & 8.3 \\
 \midrule
\multirow{9}{*}{\rotatebox{90}{Multi-Task}} & MAML & 56.4 & 54.0 & 35.0 & 28.8 & 29.1 & 29.0 & 3.0 & 4.3 & 3.7 \\
& Proto-MAML & 50.5 & 51.7 & 43.6 & 34.2 & 22.5 & 32.7 & 3.0 & \textbf{2.3} & 2.7 \\
& Meta-Curvature & 62.4 & 56.1 & 29.2 & 25.6 & 16.0 & 22.3 & 5.7 & 5.7 & 5.7 \\
& ProtoNet & 60.8 & 50.8 & 59.2 & 42.2 & 22.5 & 31.9 & \textbf{2.3} & 2.7 & \textbf{2.5} \\
& DDRR & 47.3 & 46.2 & 36.4 & 33.2 & 19.6 & 29.3 & 5.0 & 4.7 & 4.8 \\
 \cline{2-11}
& Proto-FineTuning & 47.2 & 52.1 & 33.6 & 33.9 & 19.2 & 28.1 & 6.3 & 3.7 & 5.0 \\
& FineTuning & 37.2 & 43.9 & 38.2 & 33.4 & 23.5 & 23.8 & 4.3 & 5.7 & 5.0 \\
& Linear-Readout & 40.3 & 43.7 & 24.5 & 22.3 & 21.3 & 22.8 & 7.0 & 8.0 & 7.5 \\
& TFS & 22.0 & 24.2 & 30.2 & 30.2 & 9.4 & 9.3 & 8.3 & 8.0 & 8.2 \\
\bottomrule
\end{tabular}
\caption{5-way 1-shot results, reporting the same metrics as in our primary table with variable-shot results.}\label{tab:5w1s}
\end{table*}

\begin{table*}[t]
\centering
\begin{tabular}{clccccccccc}
\toprule
&    & \multicolumn{2}{c}{\textbf{Classification}}         & \multicolumn{2}{c}{\textbf{Segmentation}}             & \multicolumn{2}{c}{\textbf{Keypoints}}                                   & \multicolumn{3}{c}{\textbf{Average Rank}}           \\
 &              & ID                & OOD                      & ID                & OOD                       & ID                & OOD                                                & ID & OOD & AVG                       \\
               \midrule
\multirow{9}{*}{\rotatebox{90}{Single-Task}} & MAML & 63.2 & 67.7 & 57.6 & 45.0 & 22.2 & 33.6 & 4.7 & 3.3 & 4.0 \\
& Proto-MAML & 57.0 & 52.3 & 49.9 & 46.7 & 22.3 & 29.6 & 5.0 & 5.7 & 5.3 \\
& Meta-Curvature & 69.0 & 67.2 & 73.5 & 53.5 & 43.7 & 16.3 & 1.7 & 4.3 & 3.0 \\
& ProtoNet & 74.3 & 64.0 & 75.9 & 55.9 & 29.4 & 33.9 & \textbf{1.3} & \textbf{2.0} & \textbf{1.7} \\
& DDRR & 68.0 & 65.7 & 69.4 & 50.5 & 22.0 & 32.0 & 4.3 & 3.7 & 4.0 \\
 \cline{2-11}
& Proto-FineTuning & 52.9 & 52.0 & 65.9 & 48.7 & 22.2 & 33.9 & 5.0 & 4.3 & 4.7 \\
& FineTuning & 43.8 & 50.0 & 55.3 & 42.7 & 22.1 & 33.1 & 6.7 & 6.3 & 6.5 \\
& Linear-Readout & 53.6 & 55.0 & 32.3 & 32.8 & 20.0 & 27.1 & 8.0 & 7.3 & 7.7 \\
& TFS & 33.8 & 44.4 & 47.4 & 40.5 & 20.9 & 28.2 & 8.3 & 8.0 & 8.2 \\
 \midrule
\multirow{9}{*}{\rotatebox{90}{Multi-Task}} & MAML & 68.3 & 72.1 & 52.0 & 42.1 & 20.7 & 31.4 & 3.3 & 3.0 & 3.2 \\
& Proto-MAML & 67.0 & 71.2 & 63.0 & 48.5 & 23.2 & 34.0 & \textbf{2.7} & 2.3 & \textbf{2.5} \\
& Meta-Curvature & 76.7 & 73.8 & 49.7 & 38.1 & 19.6 & 27.7 & 4.3 & 5.3 & 4.8 \\
& ProtoNet & 71.0 & 63.4 & 64.7 & 52.4 & 19.7 & 34.5 & 3.0 & \textbf{2.0} & \textbf{2.5} \\
& DDRR & 58.0 & 59.2 & 42.5 & 38.3 & 23.5 & 30.0 & 4.7 & 6.0 & 5.3 \\
 \cline{2-11}
& Proto-FineTuning & 52.7 & 51.3 & 50.4 & 40.6 & 21.3 & 32.5 & 4.3 & 5.0 & 4.7 \\
& FineTuning & 47.8 & 54.2 & 46.6 & 41.5 & 18.1 & 22.3 & 7.3 & 6.0 & 6.7 \\
& Linear-Readout & 48.2 & 50.9 & 45.8 & 38.4 & 19.7 & 25.5 & 6.3 & 7.0 & 6.7 \\
& TFS & 22.4 & 23.9 & 40.7 & 38.3 & 15.8 & 12.1 & 9.0 & 8.3 & 8.7 \\
\bottomrule
\end{tabular}
\caption{5-way 5-shot results, reporting the same metrics as in our primary table with variable-shot results.}\label{tab:5w5s}
\end{table*}

\end{document}